\documentclass[letterpaper]{article} 
\usepackage{aaai24}  
\usepackage{times}  
\usepackage{helvet}  
\usepackage{courier}  
\usepackage[hyphens]{url}  
\usepackage{graphicx} 
\def\UrlFont{\rm}  
\usepackage{natbib}  
\usepackage{caption} 
\usepackage{algorithm}

\usepackage{float}
\usepackage{amssymb}
\usepackage{latexsym}
\usepackage{amsmath}
\usepackage{mathtools}
\usepackage[flushleft]{threeparttable}
\usepackage{array,booktabs,makecell}
\usepackage{color}
\usepackage{soul}
\usepackage{algpseudocode}
\usepackage{enumitem}
\usepackage{comment}
\usepackage{subfig}
\usepackage[usenames,dvipsnames]{xcolor}
\usepackage{cancel}
\usepackage{fancyhdr}

\usepackage{newfloat}
\usepackage{listings}
\DeclareCaptionStyle{ruled}{labelfont=normalfont,labelsep=colon,strut=off} 
\floatstyle{ruled}
\newfloat{listing}{tb}{lst}{}
\floatname{listing}{Listing}
\usepackage{multirow}
\usepackage{colortbl}

\usepackage{xcolor}
\definecolor{deepgreen}{rgb}{0.0, 0.5, 0.0}
\definecolor{bronze}{rgb}{0.8, 0.5, 0.2}

\newcolumntype{g}{>{\columncolor{CuGray}}c}
\newcolumntype{z}{>{\columncolor{CuGray}}l}

\renewcommand{\paragraph}[1]{\noindent\textbf{#1.}\,\,}

\usepackage{xspace}

\def\onedot{.\@\xspace}
 
\def\ie{\emph{i.e}\onedot} 
\def\cf{\emph{cf}\onedot}

\def\etal{\emph{et al}\onedot}
\def\aka{\emph{a.k.a.}}

\newcommand{\Sref}[1]{Sec.~\ref{#1}}
\newcommand{\Eref}[1]{Eq.~(\ref{#1})}
\newcommand{\Fref}[1]{Fig.~\ref{#1}}
\newcommand{\Tref}[1]{Table~\ref{#1}}
\newcommand{\Aref}[1]{Alg.~\ref{#1}}

\newcommand{\be}{\begin{eqnarray}}
\newcommand{\ee}{\end{eqnarray}}
\newcommand{\bee}{\begin{eqnarray*}}
\newcommand{\eee}{\end{eqnarray*}}

\newcommand{\matrixb}{\left[ \begin{array}}
\newcommand{\matrixe}{\end{array} \right]}

\title{ContactGen: 
Contact-Guided Interactive 3D Human Generation for Partners}
\author{
    Dongjun Gu, Jaehyeok Shim, Jaehoon Jang, Changwoo Kang, Kyungdon Joo\footnote{Corresponding author.}
}
\affiliations{
    Artifcial Intelligence Graduate School, UNIST\\

    \{djku1020, jh.shim, erick1997, kangchangwoo, kyungdon\}@unist.ac.kr
}

\begin{document}

\maketitle

\begin{abstract}
Among various interactions between humans, such as eye contact and gestures, physical interactions by contact can act as an essential moment in understanding human behaviors.
Inspired by this fact, given a 3D partner human with the desired interaction label, we introduce a new task of 3D human generation in terms of physical contact.
Unlike previous works of interacting with static objects or scenes, a given partner human can have diverse poses and different contact regions according to the type of interaction.
%
%
To handle this challenge, we propose a novel method of generating interactive 3D humans for a given partner human based on a guided diffusion framework (\texttt{\small ContactGen} in short).
Specifically, we newly present a contact prediction module that adaptively estimates potential contact regions between two input humans according to the interaction label.
%
Using the estimated potential contact regions as complementary guidances, we dynamically enforce \texttt{\small ContactGen} to generate interactive 3D humans for a given partner human within a guided diffusion model. 
%
%
%
%
We demonstrate \texttt{\small ContactGen} on the CHI3D dataset, where our method generates physically plausible and diverse poses compared to comparison methods. 
%
%
Source code is available at \texttt{https://dongjunku.github.io/contactgen}.
\end{abstract}


\section{Introduction}  \label{sec:intro}

Can you imagine a scenario where a human doing any action without interacting with others?
Human actions, instead of being solely independent actions like walking and running, often stem from various interactions. 
For instance, we constantly interact with diverse objects in our surroundings and engage in daily communication with others~\cite{Yi_2023_CVPR, Hassan_2023_SIG, yi2022humanaware, Zhang_2020_CVPR}. 
Particularly, interactions among humans are vital for establishing social relationships, exchanging information, sharing emotions, and fulfilling various other aspects of life~\cite{Juulia_NAS1_2015, Rudovic_FRAI_2017, leclere_2017_TP}.

\begin{figure}[t]
    \centering
    \includegraphics[width=.95\linewidth]{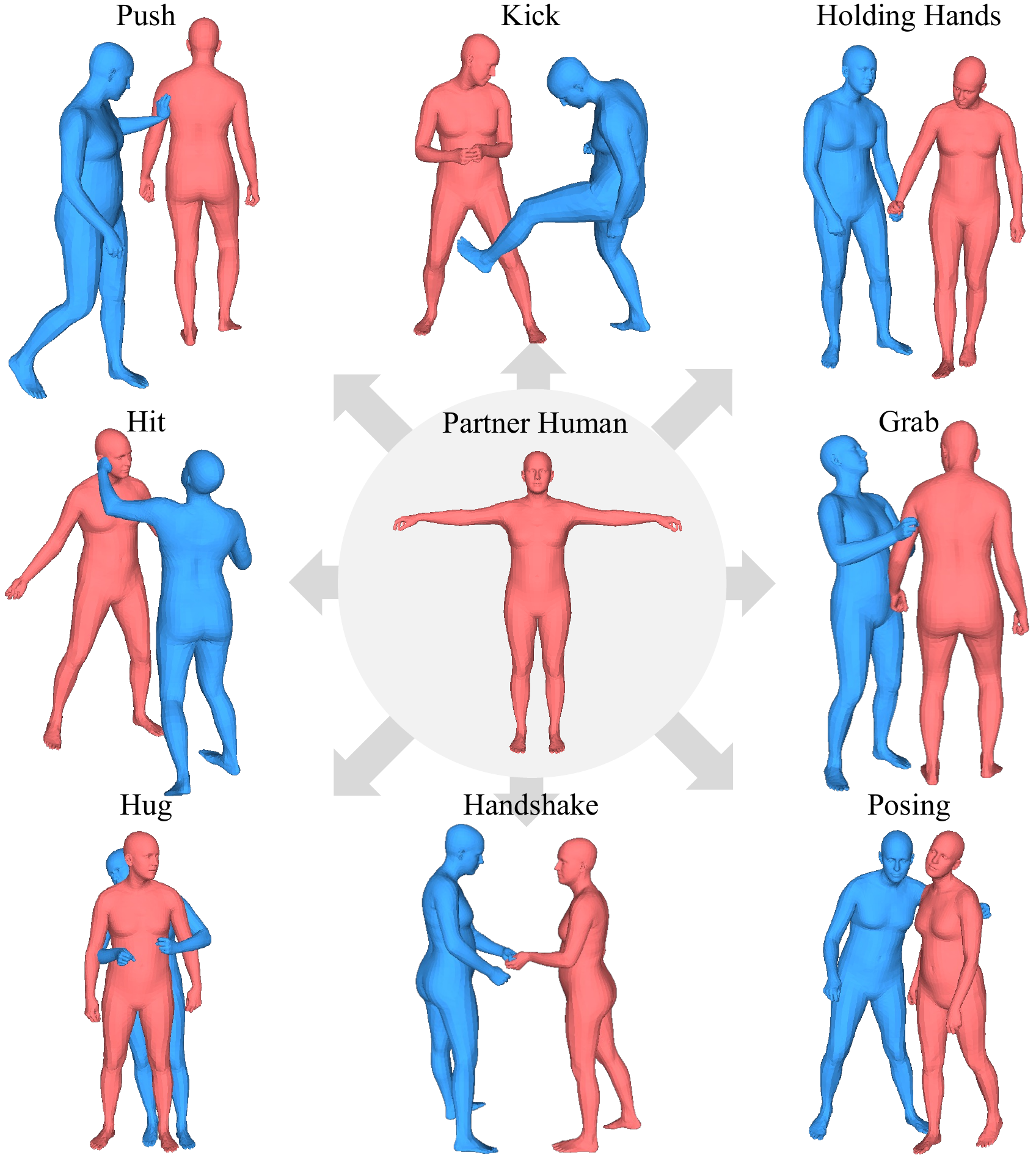}
    \vspace{-1mm}
    \caption{
        Examples of generated 3D interactive humans using the proposed \texttt{\small ContactGen}.
        Given a partner human (red-colored human) and interaction label as conditions, \texttt{ContactGen} generates various interactive 3D  humans (blue-colored human) through a guided DDM, where we focus on physical interaction by contact. 
    }
    \vspace{-5mm}
    \label{fig:teaser}
\end{figure}

Interactions among humans take on diverse forms, ranging from eye contact or gestures during conversations to physical contact. 
Among various human interactions, physical contact is common and serves as a crucial communication tool, conveying unambiguous meanings. 
Even a brief moment of physical contact can provide valuable insights into the nature of the interaction (see \Fref{fig:teaser}). 
Furthermore, it can act as a significant anchor posture for predicting human motion, as demonstrated in \cite{wu2022saga}. 
Despite the potential of interaction-based approaches, most research works in 3D human generation have concentrated on generating humans from sound~\cite{zhu2023taming_CVPR}, scene~\cite{kim2023pose}, text~\cite{petrovich2022temos, kim2023flame} and prior movement inputs~\cite{Lee_Moon_Lee_2023}, overlooking the aspect of the interaction between humans.
Instead, physical interactions are primarily addressed in hand-object interaction~\cite{Grady_2021_CVPR,Karunratanakul2020GraspingFL} and scene-based interaction studies~\cite{Zhang_2020_CVPR, hassan2021populating}.

In this work, we investigate a new problem of a 3D human generation that maintains natural physical contact with a given partner human based on input interaction labels.
Unlike previous works on interactions between humans and objects (or scenes)~\cite{ zhao2022compositional, Huang_2023_CVPR, Yi_2023_CVPR}, generating 3D human interacting with a given partner human involves several challenges.
Specifically, the target of interaction (\ie, partner humans) is non-rigid and dynamic, leading to numerous possible combinations of pose and shape.
In addition, compared to objects (or scenes) that have a clearly defined purpose, humans can have different physical contact regions based on the type of interactions.
For instance, when two humans interact, such as a greeting, the points of contact can exhibit considerable variation.
They might opt for a handshake, a hug, or even a cheek-to-cheek kiss.
Each of these interactions has different physical contact regions, making the interaction more complex and dynamic than with static objects or scenes.

To overcome these challenges, we present a novel approach to generate an interactive 3D human that maintains natural physical contact with a given partner human according to the desired interaction label.
%
We call this proposed method \texttt{\small ContactGen} (see \Fref{fig:overview}).
\texttt{\small ContactGen} basically adopts a guided denoising diffusion model (DDM)~\cite{dhariwal2021diffusion,Huang_2023_CVPR}, which enables to consider additional optimization as guidance during the sampling process of DDM.
%
In particular, we aim to exploit physical interaction, such as contact, as guidance within a guided DDM.
To this end, we propose a contact prediction module that adaptively estimates the probability of physical contact for each body part between two input humans according to the interaction label. 
Based on the predicted contact regions, we can seamlessly incorporate interactive optimization related to physical contact into DDM, which allows us to synthesize interactions that involve meaningful and accurate points of contact, enhancing the realism and coherence of the generated human interactions.
We demonstrate the effectiveness and realism of our method on the modified version of the CHI3D dataset~\cite{fieraru2020three}, where \texttt{\small ContactGen} shows a promising solution for generating rich and dynamic human interactions. 

In summary, our main contributions are:
\begin{itemize}
\item We introduce a new and exciting task of generating interactive 3D humans for a given partner human and desired interaction label. 
%
\item We propose a novel guided DDM for generating interactive 3D humans, \texttt{\small ContactGen}, that considers physical interactions, such as contact during sampling, leading to physically feasible human interactions.
\item We present a contact prediction module that adaptively predicts potential interaction regions between humans, which allows us to optimize valid interaction regions within a guided DDM.
\end{itemize}

\section{Related Work}  \label{sec:related_work}
%
Our method of generating physical interactions by contact between humans has not been studied before, but related studies do exist. 
To identify our work, we summarize relevant research on human action generation, physics-based prediction, and the human-to-human interaction dataset.

%


\subsubsection{Human action generation.}
%
Since human action recognition and prediction are crucial for understanding human behavior, many studies have been actively conducted in the computer vision field~\cite{Tran_2018_CVPR, Shi_2019_CVPR, Wang_2021_CVPR, Duan_2022_CVPR}.
Inspired by the success of the generation model in the 2D domain, many works have been proposed to synthesize 3D human motion from various modalities.
Given an action label, Action2Motion~\cite{guo2020action2motion} and ACTOR~\cite{petrovich2021action} adopt a variational autoencoder (VAE) architecture ~\cite{kingma2013auto} that learns latent space representing human motions.
Beyond this simple conditioning, TEMOS~\cite{petrovich2022temos} and TEACH~\cite{athanasiou2022teach} take text descriptions that contain multiple actions as input, creating a natural motion sequence.
However, these VAE-based approaches, which are prone to posterior collapse, have limitations of human motion diversity. \\
%
\indent To alleviate this issue, Tevet \etal~\cite{tevet2023human} propose MDM, the first time to apply a diffusion model~\cite{ho2020denoising} to human motion synthesis.
%
PhysDiff~\cite{yuan2022physdiff} extends MDM to enforce the generated motion sample to be physically plausible by projecting physical constraints from the physics simulator to the diffusion process.
Although the above methods show great performance, they only focus on synthesizing a single human motion, not considering interaction between humans.
On the other hand, we propose a diffusion-based framework for generating a 3D human who comes into contact with various body parts of a given partner based on input interaction.

\subsubsection{Physics-based prediction in 3D.}
Physical contact-based interactions are most extensively covered in scene-aware 3D human generation~\cite{Zhang_2020_CVPR, zhao2022compositional, Huang_2023_CVPR}, object-aware 3D hand pose estimation~\cite{taheri2022goal, wu2022saga}, and motion re-targeting~\cite{van2016spatio, mehta2018single, guo2022multi}.
Several research works use proximity between human and a given scene to induce physical affordance and use negative SDF to prohibit interpenetration~\cite{Zhang_2020_CVPR}. 
Other methods use object alignment by point normals~\cite{kim2023pose,Grady_2021_CVPR}, and point-wise embedded closest semantics~\cite{kim2023pose}.
In contrast to previous studies, human-human interactions are challenging due to their dynamic and non-rigid nature.
As a result, research on human-human interaction is limited to the subject of 3D motion re-targeting and is given relatively little attention.
To handle this challenge, we propose a contact prediction module that can predict potential contact regions between each human body during the sampling process and provide guidance for diffusion.
%

\subsubsection{Dataset of human-human interaction in 3D.}
Contrary to many studies~\cite{van2016spatio, mehta2018single, guo2022multi} that introduce 3D datasets addressing human interactions, they represented the 3D human solely based on joint locations, omitting contact annotations.
To better understand human interactions in 3D, Fieraru~\etal~\cite{fieraru2020three} provide the CHI3D and FlickrCI3D datasets that contain region-level contact information with human body meshes.
Furthermore, Yin~\etal~\cite{yin2023hi4d} introduce Hi4D, which additionally includes 4D textured scans with interaction-centric annotations at the body vertex.
Exploiting their abundant contact information, BUDDI~\cite{mueller2023buddi} reconstructs interacting humans from images and then applies the diffusion process on the SMPL-X parameters~\cite{pavlakos2019expressive} to refine reconstruction quality.
Although this method achieves high reconstruction performance of two humans, it requires the off-the-shelf network to initialize SMPL-X.
Contrarily, \texttt{\small ContactGen} generates diverse interactive 3D humans corresponding to given interaction labels without any additional network or initialization.


\section{Methodology}  \label{sec:method}

\begin{figure}[t]
    \centering
    \includegraphics[width=.90\linewidth]{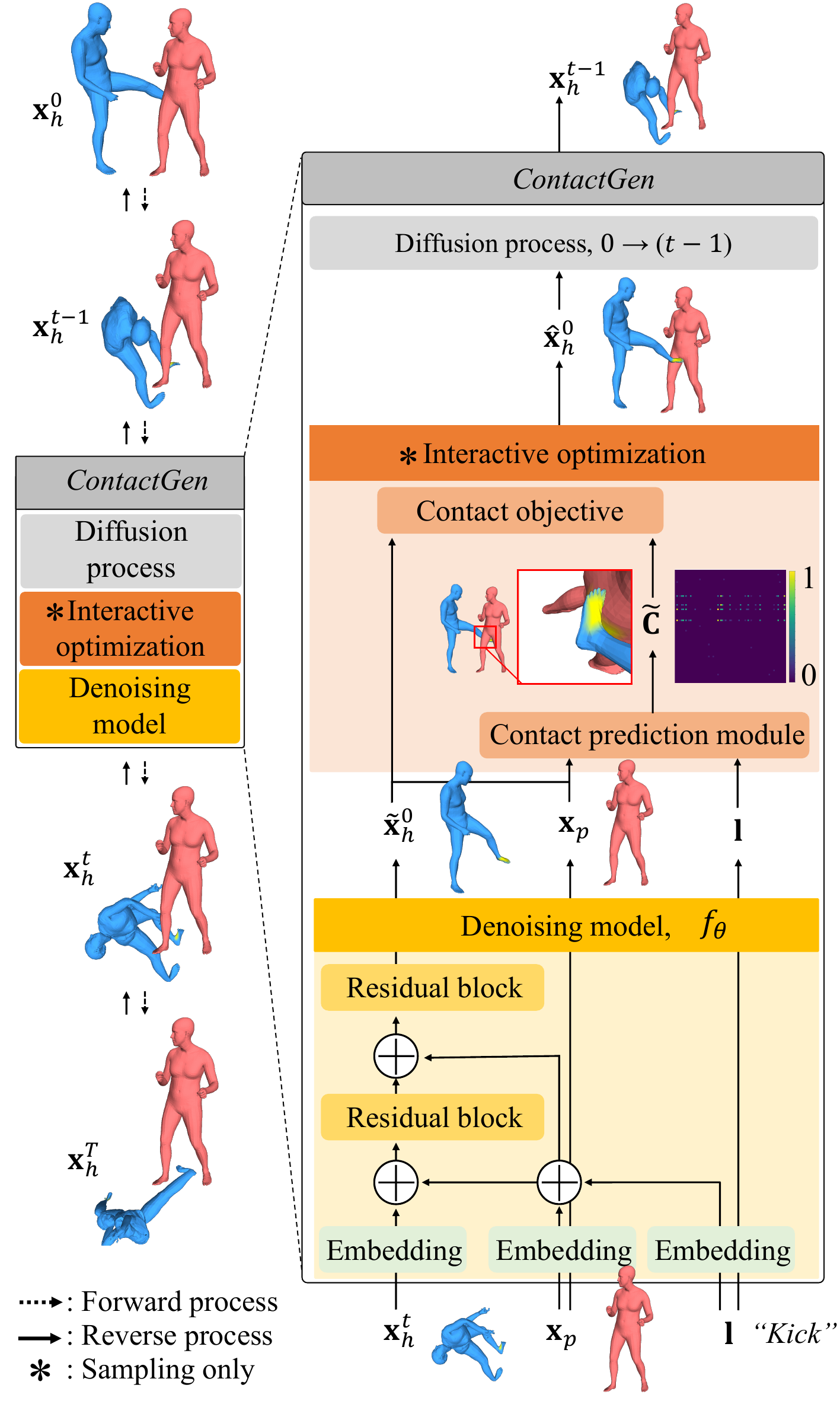}
    \vspace{-2mm}
    \caption{
        Overview of the \texttt{ContactGen}. 
        Given a partner $\mathbf{x}_p$ and interaction label $\mathbf{l}$, our method generates an interactive 3D human $\mathbf{x}_h^0$ using a guided DDM.
        In particular, we adaptively estimate potential contact regions between humans by the contact prediction module (orange box). Based on the potential contact regions, we perform interactive optimization by considering physical contact, which provides additional guidance during the sampling.
    \vspace{-2.5mm}
    }
    \label{fig:overview}
\end{figure}

In this work, we propose a new method to generate an interactive 3D human that maintains natural physical contact with a given partner human according to the desired interaction label; we call the proposed approach \texttt{\small ContactGen} (see the overall architecture in \Fref{fig:overview}).

Concretely, \texttt{\small ContactGen} takes as input a partner human $\mathbf{x}_p$, represented by SMPL-X~\cite{pavlakos2019expressive}, and the desired interaction label~$\mathbf{l}$ as conditions.
\texttt{\small ContactGen} then aims to generate the corresponding interactive 3D human $\mathbf{x}_h$ in terms of physical contact, such as kick, grab, and hug. 
To this end, \texttt{\small ContactGen} basically adopts a guided DDM~\cite{dhariwal2021diffusion} that can integrate optimization as guidance into each step of the sampling process. 
Within a guided DDM, \texttt{\small ContactGen} proposes a part-wise contact prediction that estimates potential contact regions for two input humans and target interaction, which allows us to perform contact-aware optimization and generate physically plausible humans in terms of contact.

\subsection{Guided DDM for Interactive 3D Human}  \label{subsec:gddm} 

Guided denoising diffusion (\aka~guided sampling) is a variant of the diffusion sampling process~\cite{dhariwal2021diffusion}. 
Guided sampling introduces an optimization step performed after the denoising step in each iteration of the sampling process, which allows us to sample data satisfying the specified condition represented as an objective function from the pre-estimated distribution of DDM.
Based on this fact, we adopt a guided DDM for interactive 3D human generation in terms of physical contact.
Specifically, we use a conditional DDM as the backbone to consider a given partner and interaction label and then expand its sampling process to a guided sampling process to take into account physical contact effectively.


%
%
%
%

\subsubsection{Forward process.}
%
In the forward (diffusion) process, data $\mathbf{x}_h^0$ (\ie,~interactive 3D human in our case) 
is gradually corrupted by Gaussian noise, ultimately becoming the standard Gaussian distribution. 
This diffusion process can be represented as a Markov chain: \vspace{-2mm}
\begin{equation}
    q(\mathbf{x}_h^{0:T}) = q(\mathbf{x}_h^0)\prod_{t=1}^T q(\mathbf{x}_h^{t}|\mathbf{x}_h^{t-1}),
\end{equation}
\begin{equation}
    q(\mathbf{x}_h^{t}|\mathbf{x}_h^{t-1}) = \mathcal{N}(\mathbf{x}_h^{t};\sqrt{1 - \beta_t}\mathbf{x}_h^{t-1}, \beta_{t}\mathbf{I}),
\end{equation}
where $\beta_t$ denotes the noise scheduling parameter.
%
According to \cite{ho2020denoising}, this forward process admits sampling $\mathbf{x}_h^t$ at an arbitrary time step $t$ by a closed form:
%
\begin{equation}
    q(\mathbf{x}_h^t|\mathbf{x}_h^0) = \mathcal{N}(\mathbf{x}_h^t;\sqrt{\hat{\alpha}_t}\mathbf{x}_h^0, (1 - \hat{\alpha}_t)\mathbf{I}),
\end{equation}
where $\alpha_t = 1 - \beta_t$ and $\hat{\alpha}_t = \prod_{i=1}^t\alpha_i$. 
%

\subsubsection{Conditional reverse process.} 
%
In the conditional reverse process, we take a partner human $\mathbf{x}_p$ and interaction label $\mathbf{l}$ as conditions, then gradually restore an interactive human $\mathbf{x}_h^t$ by predicting noise. 
This conditional process also can be formulated using a Markov chain: \vspace{-2mm}
\begin{equation}
    p_\theta(\mathbf{x}_h^{0:T}|\mathbf{x}_p, \mathbf{l}) = p(\mathbf{x}_h^T)\prod_{t=1}^T p_\theta(\mathbf{x}_h^{t-1}|\mathbf{x}_h^t, \mathbf{x}_p, \mathbf{l}),
    \label{eqn:reverse}
\end{equation}
\begin{equation}
    p_\theta(\mathbf{x}_h^{t-1}|\mathbf{x}_h^t, \mathbf{x}_p, \mathbf{l}) = \mathcal{N}(\mathbf{x}_h^{t-1};\mu_\theta(\mathbf{x}_h^t, t, \mathbf{x}_p, \mathbf{l}), \sigma_t^2\mathbf{I}),
    \label{eqn:generation}
\end{equation}
where $p(\mathbf{x}_h^T)$ is the standard Gaussian distribution, 
$\mu_\theta(\mathbf{x}_h^t, t, \mathbf{x}_p, \mathbf{l})$ is the mean induced from the noise prediction network $\epsilon_\theta(\mathbf{x}_h^t, t, \mathbf{x}_p, \mathbf{l})$ parameterized by $\theta$, 
and $\sigma_t$ is a time-step dependant variance.
%
By repeating this process, we can eventually generate an interactive 3D human $\mathbf{x}_h^0$.

\subsubsection{Training objective.} 
%
Noise prediction network $\epsilon_\theta$ is trained with MSE between predicted and ground truth noises: \vspace{-1mm}
\begin{equation}
    \mathcal{L}(\tilde{\epsilon}, \epsilon)=\Sigma_{i} \frac{1}{N_i} \left\| \tilde{\mathbf{\epsilon}}_i - \mathbf{\epsilon}_i \right\|^2_2,
\end{equation}
where $\tilde{\epsilon}_i$ and $\epsilon_i$ indicate parts of predicted noise $\tilde{\epsilon}=\epsilon_\theta(\mathbf{x}_h^t, t, \mathbf{x}_p, \mathbf{l})$ and ground truth noise $\epsilon$, respectively, and $N_i$ means the dimension of $\epsilon_i$.
We split the predicted noise $\tilde{\epsilon}$ into four parts of $\tilde{\epsilon}_i$ according to their characteristics, such as translation, rotation, body pose, and hand poses.
By dividing, we aim to lead our model focusing on each human characteristic with the same weight.
%
The detailed procedure for training conditional DDM is available in \Aref{alg:train}.

\newcommand\Algphase[1]{%
\vspace{-.7\baselineskip}\Statex\hspace{\dimexpr-\algorithmicindent-2pt\relax}\rule{\textwidth}{0.4pt}%
\Statex\hspace{-\algorithmicindent}\textbf{#1}%
\vspace{-.7\baselineskip}\Statex\hspace{\dimexpr-\algorithmicindent-2pt\relax}\rule{\textwidth}{0.4pt}%
}
\renewcommand{\algorithmicrequire}{\textbf{Input:}}
\renewcommand{\algorithmicensure}{\textbf{Output:}}
\begin{algorithm}[t]
\caption{Training}\label{alg:train}
{\footnotesize
\begin{algorithmic}
    \Require learning rate $\eta$
    \Repeat    
    \State $(\mathbf{x}_h^0, \mathbf{x}_p, \mathbf{l}) \sim q(\mathbf{x}_h^0, \mathbf{x}_p, \mathbf{l})$
    \State $t \sim \mathcal U(\{1, \dots T\})$
    \State $\mathbf{\epsilon} \sim \mathcal{N}(\mathbf{0},\mathbf{I})$
    \State $\mathbf{x}_h^t = \sqrt{\hat{\alpha}_t}\mathbf{x}_h^0 {+} \sqrt{1-\hat{\alpha}_t}\epsilon$
    \State $\tilde{\epsilon} = \epsilon_\theta(\mathbf{x}_h^t, t, \mathbf{x}_p, \mathbf{l})$
        \Comment{noise prediction}
    \State $\theta \leftarrow \eta\nabla_\theta \mathcal{L}(\tilde{\epsilon}, \epsilon)$
\Until{$\epsilon_\theta$ is converged}
\end{algorithmic}
}
\end{algorithm}

\subsubsection{Guided sampling.} \ 
%
Beyond the conditional sampling from $p_\theta(\mathbf{x}_h^{0:T}|\mathbf{x}_p, \mathbf{l})$ by \Eref{eqn:reverse}, we introduce a guided sampling that allows us to sample from $p_\theta(\mathbf{x}_h^{0:T}|\mathbf{x}_p, \mathbf{l}, \mathcal{O})$, while satisfying a certain objective $\mathcal{O}$ with the learned $p_\theta(\mathbf{x}_h^{0:T}|\mathbf{x}_p, \mathbf{l})$.
%
%

First of all, the distribution $p_\theta(\mathbf{x}_h^{0:T}|\mathbf{x}_p, \mathbf{l}, \mathcal{O})$ we desire to sample can be formulated using the form of a Markov chain: \vspace{-1mm}
\begin{equation}
    \small
    \label{eqn:ddim_sampling1}
    p_\theta(\mathbf{x}_h^{0:T}|\mathbf{x}_p, \mathbf{l}, \mathcal{O}) = p(\mathbf{x}_h^T)\prod_{t=1}^T p_\theta(\mathbf{x}_h^{t-1}|\mathbf{x}_h^t, \mathbf{x}_p, \mathbf{l}, \mathcal{O}).
\end{equation}
%
%
Here, we use the reverse process $q'(\mathbf{x}_h^{t-1} | \mathbf{x}_h^t, \mathbf{x}_h^0)$ defined in denoising diffusion implicit models (DDIM) \cite{song2020denoising}. 
In each iteration of this reverse process, the model predicts $\mathbf{x}_h^{0}$ from a noisy input $\mathbf{x}_h^{t}$ and the predicted noise: 
\begin{equation}
    \small
    \label{eqn:human_denoising}
    f_\theta(\mathbf{x}_h^t, t, \mathbf{x}_p, \mathbf{l}) = (\mathbf{x}_h^t - \sqrt{1 - \hat{\alpha}_t} \epsilon_\theta(\mathbf{x}_h^t, t, \mathbf{x}_p, \mathbf{l})) / \sqrt{\hat{\alpha}_t}.
\end{equation}
Then, the model subsequently generates a sample $\mathbf{x}_h^{t-1}$ using the reverse process $q'(\mathbf{x}_h^{t-1} | \mathbf{x}_h^t, \mathbf{x}_h^0)$.

Concretely, $p_\theta(\mathbf{x}_h^{t-1}|\mathbf{x}_h^t, \mathbf{x}_p, \mathbf{l}, \mathcal{O})$ can be represented as: 
%
\begin{equation}
    \small
    \label{eqn:ddim_sampling}
    p_\theta(\mathbf{x}_h^{t-1}|\mathbf{x}_h^t, \mathbf{x}_p, \mathbf{l}, \mathcal{O}) = q'(\mathbf{x}_h^{t-1} | \mathbf{x}_h^t, f_\theta(\mathbf{x}_h^t, t, \mathbf{x}_p, \mathbf{l}, \mathcal{O})),
\end{equation}
where $f_\theta(\mathbf{x}_h^t, t, \mathbf{x}_p, \mathbf{l}, \mathcal{O})$ indicates objective-aware denoised human prediction function based on $f_\theta(\mathbf{x}_h^t, t, \mathbf{x}_p, \mathbf{l})$ and $\mathcal{O}$. 
To consider a certain objective $\mathcal{O}$, we introduce guidance as:
\begin{equation}
    \label{eqn:obj}
    \mathbf{g} = \nabla_{\mathbf{x}_h}\mathcal{O}(\mathbf{x}_h, \mathbf{x}_p, \mathbf{l})|_{\mathbf{x}_h=f_\theta(\mathbf{x}_h^t, t, \mathbf{x}_p, \mathbf{l})}.
\end{equation}
Then, we define $f_\theta(\mathbf{x}_h^t, t, \mathbf{x}_p, \mathbf{l}, \mathcal{O})$ as:
\begin{equation}
    \label{eqn:human_denoising_obj}
    f_\theta(\mathbf{x}_h^t, t, \mathbf{x}_p, \mathbf{l}, \mathcal{O}) = f_\theta(\mathbf{x}_h^t, t, \mathbf{x}_p, \mathbf{l}) - \mathbf{\lambda}\odot\mathbf{g},
\end{equation}
where $\mathbf{\lambda}$ denotes the scaling vector for the guidance and $\odot$ indicates the element-wise product.
In \Eref{eqn:human_denoising_obj}, we apply the guidance $\mathbf{g}$ in a compact form for iterative computation.

In this work, we consider physical contact between humans as the objective $\mathcal{O}$ to provide guidance (\cf \Sref{subsec:opt} for details of guidance objective).
Specifically, in each iteration of the reverse process described above, we estimate noise-less human $\tilde{\mathbf{x}}_h^0$ from the estimated noise $\tilde{\epsilon}$ and a noisy input $\mathbf{x}_h^t$ in \Eref{eqn:human_denoising}.
Then, $\tilde{\mathbf{x}}_h^0$ is further optimized to satisfy the objective $\mathcal{O}$, leading to optimized human $\hat{\mathbf{x}}^0_h$ in \Eref{eqn:human_denoising_obj}.

Note that instead of directly optimizing $\mathbf{x}_h^{t - 1}$, as in \cite{Huang_2023_CVPR}, \texttt{\small ContactGen} indirectly optimizes $\mathbf{x}_h^{t - 1}$ by optimizing $\tilde{\mathbf{x}}_h^0$ with the DDIM sampling process.
As a result, \texttt{\small ContactGen} can effectively estimate contact regions from less noisy humans $\tilde{\mathbf{x}}_h^0$ instead of noisy ones $\mathbf{x}_h^{t - 1}$.
The detailed guided sampling algorithm is summarized in \Aref{alg:sample}. 

\begin{figure}
    \centering
    \includegraphics[width=.95\linewidth]{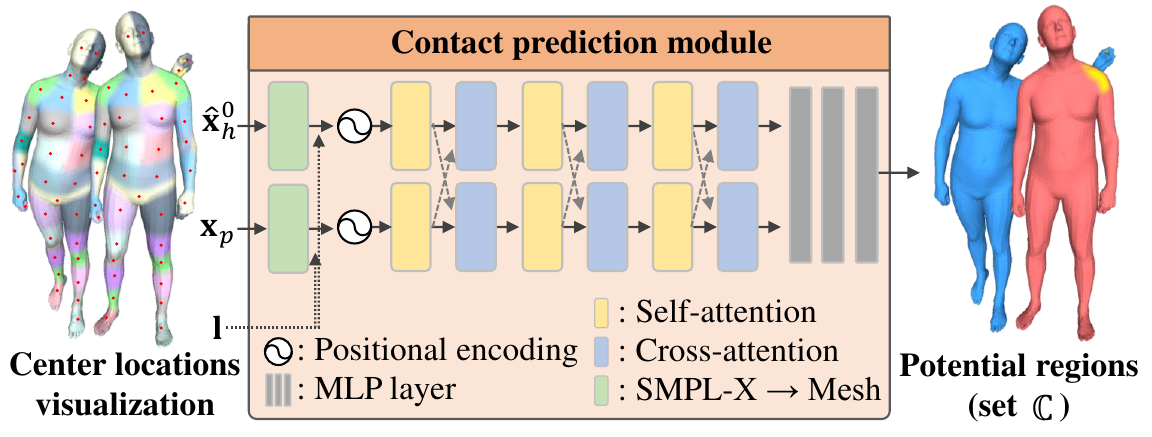}
    \vspace{-1mm}
    \caption{
        Illustration of contact prediction module.
        We use the contact prediction module to predict the part-wise contact probability map $\mathbf{C}$ between interacting humans.
        After estimating $\mathbf{C}$, a set of potential contact regions $\mathbb{C}$ (right) can be obtained from a certain threshold $\tau$.
    }
    \label{fig:contact}
\end{figure}

\begin{algorithm}[t]
\caption{Guided Sampling} \label{alg:sample}
{\footnotesize
\begin{algorithmic}
    \Require partner 3D human $\mathbf{x}_p$, interaction label $\mathbf{l}$
    \Ensure generated interactive 3D human $\mathbf{x}_h^{t-1}$
    \State $\mathbf{x}_h^T \sim \mathcal{N}(\mathbf{0},\mathbf{I})$
    \For{$t = T, ..., 1$}
        \State $\tilde{\mathbf{x}}_h^0 = f_\theta(\mathbf{x}_h^t, t, \mathbf{x}_p, \mathbf{l})$
            \Comment{denoised sample by \Eref{eqn:human_denoising}}
        \State $\mathbf{g} = \nabla_{\mathbf{x}_h}\mathcal{O}(\mathbf{x}_h, \mathbf{x}_p, \mathbf{l})|_{\mathbf{x}_h=\tilde{\mathbf{x}}_h^0}$
            \Comment{guidance by \Eref{eqn:obj}}
        \State $\hat{\mathbf{x}}_h^0 = \tilde{\mathbf{x}}_h^0 - \lambda\odot\mathbf{g}$
            \Comment{interactive optimization by \Eref{eqn:human_denoising_obj}}
        %
        \State $\mathbf{x}_h^{t-1} = \sqrt{\hat{\alpha}_{t-1}}\hat{\mathbf{x}}_h^0 + \frac{\sqrt{1{-}\hat{\alpha}_{t-1}}}{\sqrt{1-\hat{\alpha}_t}}\left(\mathbf{x}_h^t - \sqrt{\hat{\alpha}_t}\hat{\mathbf{x}}_h^0\right)$ \\
            \Comment{reverse process of DDIM} 
    \EndFor
\end{algorithmic}
}
\end{algorithm}

%

\subsection{Interactive Optimization}   \label{subsec:opt} 

We design optimization objectives that help the diffusion process by providing a guiding direction in terms of physical interaction by contact.

\subsubsection{Contact objective.} \
We exploit the contact objective to promote physical contact between a partner human and an interactive human so that their interaction can be physically natural.
The contact objective is formulated as:
\begin{equation}
    \mathcal{O}_{contact}(\mathbf{x}_h, \mathbf{x}_p, \mathbf{l}) = \Sigma_{(r_i, r_j) \in \mathbb{C}} \ d_{CD}(V_h^{r_i}, V_p^{r_j}),
\end{equation}
where $d_{CD}(\cdot,\cdot)$ is the Chamfer distance between two sets of vertices, $\mathbb{C}$ denotes the contact region set predicted by the contact prediction module (\cf \Sref{subsec:cpm}), and $V_h^{r_i}$ and $V_p^{r_j}$ are the set of vertices of interactive human and partner human, belonging to regions $r_i$ and $r_j$, respectively.



\subsubsection{Interactive optimization.} 
Based on the above objective, we expand $\mathcal{O}(\mathbf{x}_h, \mathbf{x}_p, \mathbf{l})$ in \Eref{eqn:obj} into customized objectives for interaction, following \cite{Huang_2023_CVPR}:
\begin{equation}
    \mathcal{O}(\mathbf{x}_h, \mathbf{x}_p, \mathbf{l}) = \mathcal{O}_{contact}(\mathbf{x}_h, \mathbf{x}_p, \mathbf{l}).
\end{equation}
\subsection{Contact Prediction Module} \label{subsec:cpm} 
To adaptively provide potential contact regions within the guided DDM, we propose a new transformer-based contact prediction module that estimates the contact probability map between each contact region of a given partner human $\mathbf{x}_p$ and an interactive human $\tilde{\mathbf{x}}^0_h$ (see \Fref{fig:contact}).
%

To this end, we divide the SMPL-X mesh of humans into $N_\text{reg}$ number of regions {with regard to vertex indices following CI3D~\cite{fieraru2020three} and predict a part-wise contact map $\mathbf{C} {\in} \{0, 1\}^{N_\text{reg} {\times} N_\text{reg}}$  (we set $N_\text{reg}$ as 75).
%
%
%
Concretely, the contact prediction module takes as input each center location of $N_\text{reg}$ regions 
of partner and interactive humans and passes through a sequence of transformers~\cite{vaswani2017attention} composed of self-attention and cross-attention to compute each feature vector. 
%
Then, we estimate part-wise contact probability map $\mathbf{C}$ using our contact prediction module by mutually extracting features from both SMPL-X parameters of the partner and the denoised interactive human.}
%
%
The contact prediction module is trained with binary-cross-entropy objective function:
\begin{equation}
    \small
    \mathcal{L}_\text{CE}{=}-\frac{1}{N_\text{reg}^2}\sum_i^{N_\text{reg}}\sum_j^{N_\text{reg}}\left(
        \mathbf{C}_{i,j}\log(\tilde{\mathbf{C}}_{i,j}) {+} (1{-}\mathbf{C}_{i,j})\log(1{-}\tilde{\mathbf{C}}_{i,j})
    \right),
\end{equation}
%
where $\mathbf{C}_{i,j}$ indicates the probability of contact occurring between $i$-th and $j$-th regions from interactive human and partner, respectively, and $\mathbf{C}$ and $\tilde{\mathbf{C}}$ are the ground truth and estimated contact prediction maps, respectively.

After estimating $\tilde{\mathbf{C}}$, we compute a contact region set  $\mathbb{C}$ that contains a set of contact region pairs corresponding to probability over a certain threshold $\tau$ (we set $\tau$ as 0.5).
By predicting 
$\mathbb{C}$ for $\tilde{\mathbf{x}}_h^0$ and ${\mathbf{x}}_p$ at each time $t$ in an adaptive way, we can provide dynamic contact regions within the guided DDM.
That is, \texttt{\small ContactGen} can sample diverse interactive humans of high quality by avoiding the conflict between diffusion direction and guiding direction.

\begin{figure*}[t]
\centering
\includegraphics[width=0.93\linewidth]{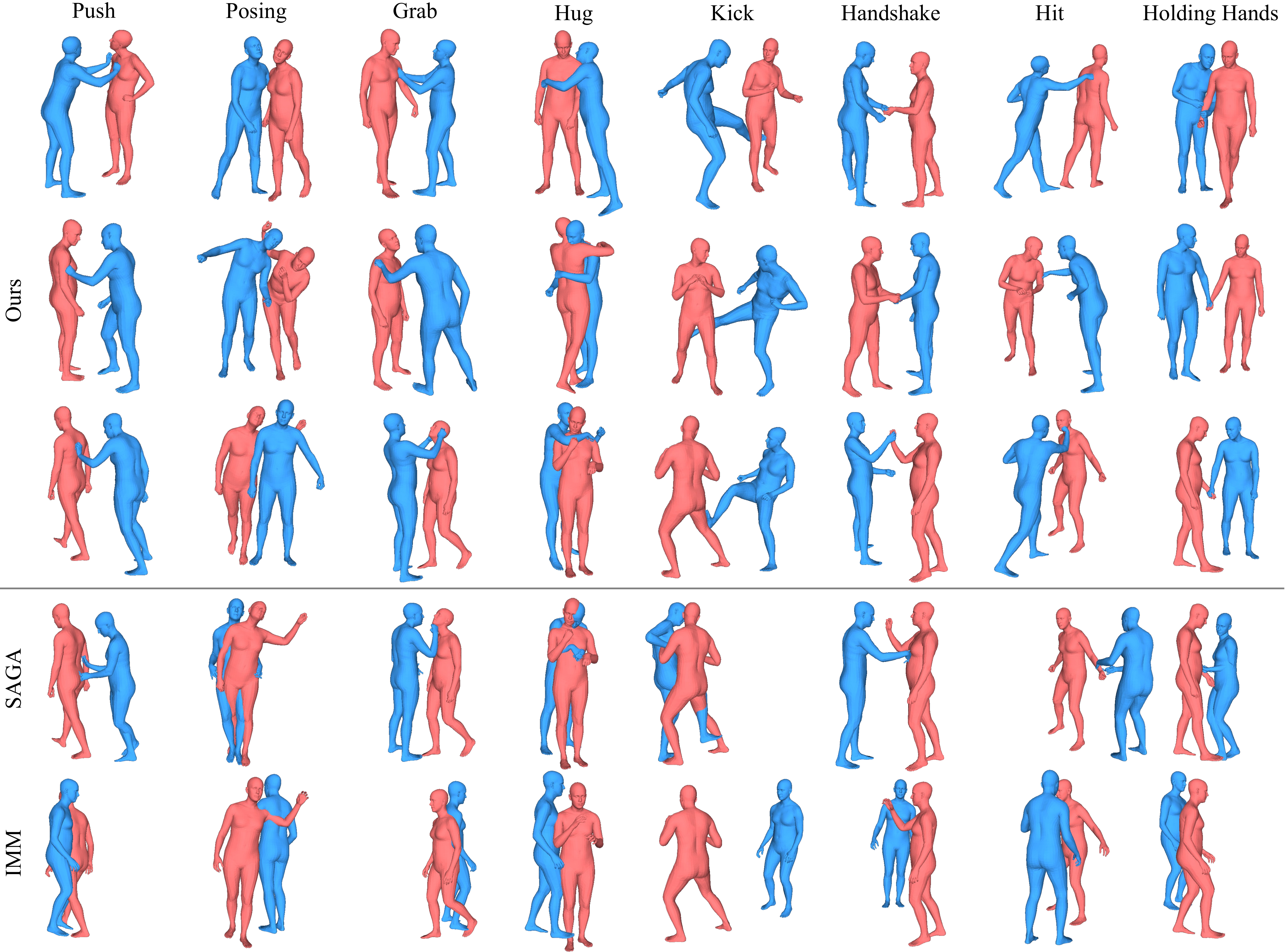}
\vspace{-1mm}
\caption{
    Qualitative evaluation {with comparison methods} on the modified CHI3D dataset.
    We visualize generated interactive humans (blue-colored humans) alongside their given partners (red-colored humans).
    The first three rows show generated humans by \texttt{ContactGen} according to different partner poses, where \texttt{ContactGen} shows diverse and plausible generations in terms of physical contact.
    %
    For comparison with SAGA and IMM, we visualize generated humans under the same given partner conditions (last three rows).
    Compared to our results showing visually satisfactory outcomes, humans generated by SAGA and IMM show improper pose and penetration.
}
\vspace{-3mm}
\label{fig:qualitative}
\end{figure*}

\begin{figure}[t]
    \centering
    \includegraphics[width=.865\linewidth]{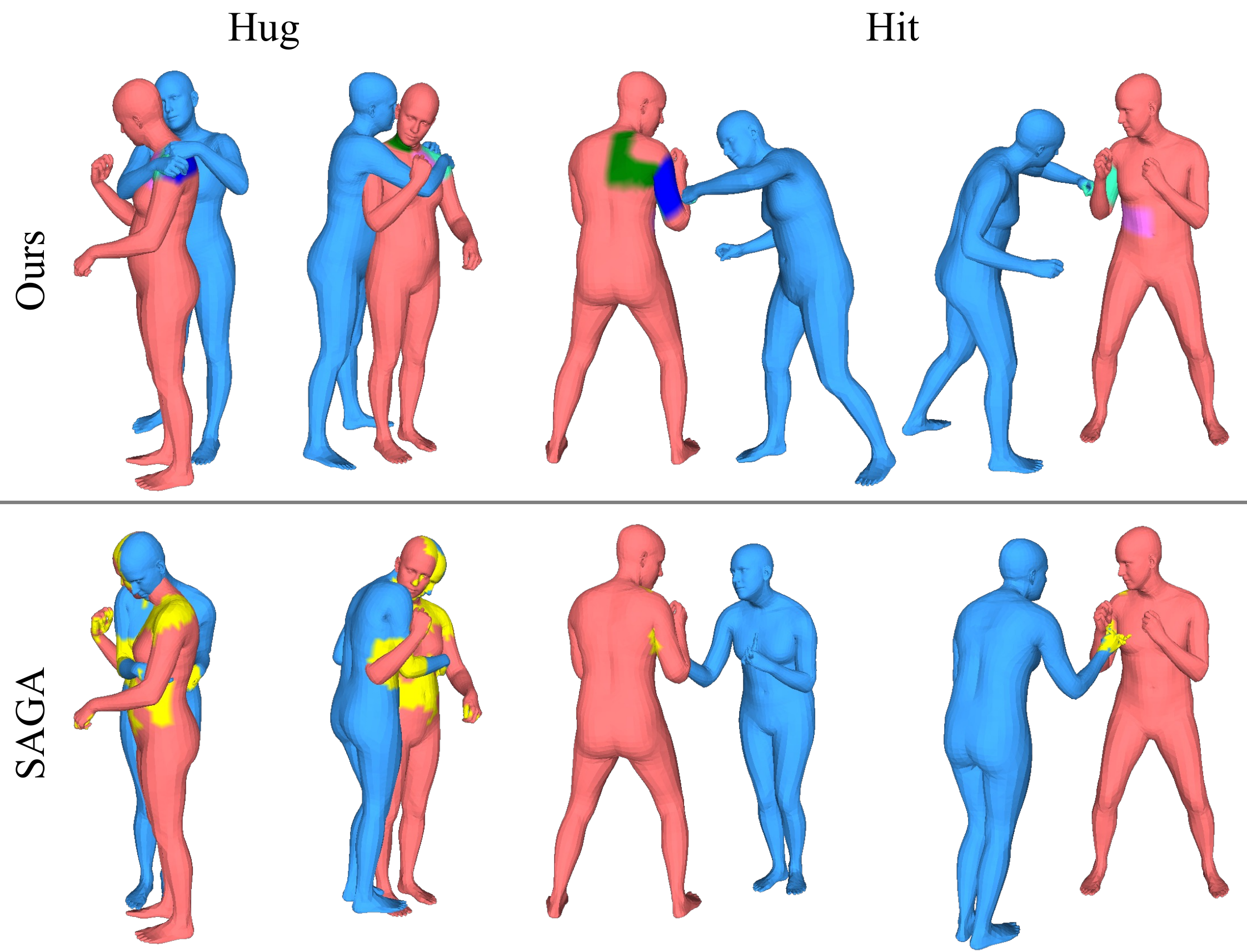}
    \vspace{-2mm}
    \caption{
        Comparison of predicted contact regions. %
        In our method, the same colored regions describe the estimated contact regions between two humans. 
        We can observe plausible contact regions according to the interaction label.
        In SAGA, yellow indicates the estimated contactable vertices between humans, but they do not know the correspondences.
        %
    }
    \label{fig:contact_red}
\end{figure}

\section{Experiments}  \label{sec:experiment}

\subsection{Implementation Details} 
The training policy of our guided DDM basically follows DDM~\cite{ho2020denoising}.
%
%
We selectively adopt SMPL-X representation~\cite{pavlakos2019expressive}, a differentiable function that maps from a set of low-dimensional body parameters to a 3D human mesh.
We use the Adam optimizer~\cite{KingBa15} to train our networks.
For the training procedure, noise is sampled from the linear noise scheduler initialized with $\beta_0 =5e-6$, $\beta_T = 5e-3$, and diffusion step $T = 1000$.
In the sampling phase, we utilize classifier-free guidance \cite{ho2022classifier} by hiding partner and both partner and interaction label conditions in the training and sampling of our network.
%
%
More details of implementation are described in supplementary materials.

\subsection{Dataset} \label{subsec:dataset}
We basically use CHI3D~\cite{fieraru2020three}, which is a 3D motion capture dataset of 8 close human interaction scenarios: push, posing, grab, hug, kick, handshake, holding hands and hit.
For each interaction, CHI3D provides various sequences (\ie, video, not contact interaction moment) containing physical contact events and corresponding 3D human models, such as SMPL-X~\cite{pavlakos2019expressive} and GHUM~\cite{xu2020ghum}.
To generate a 3D human in contact with a given partner based on input interaction, we proceed with several pre-processing on CHI3D to acquire a valid dataset for training interactive human generation in terms of physical contact.


Concretely, we perform frame sampling that captures the moment of contact between human subjects.
We then normalize the rotation parameters so that all partners head in the same direction, and transform the rotation and translation parameters of the actor to maintain the interaction.
In addition, we augment the 3D human representation in terms of the body parts of a human with symmetry to ensure the diversity of generation results.
For example, we flip the two people holding their right hands so that they hold their left hands.
We also utilize contact annotation between human subjects to train the contact prediction module.
%
We will release additional code to preprocess CHI3D dataset for research purposes.

\subsection{Evaluation Metrics} \label{subsec:eval}
We evaluate and compare \texttt{\small ContactGen} in terms of sample quality and physical plausibility.
To this end, we propose Fréchet human interaction distance~(FHID) to assess sample quality by comparing distributional discrepancy between a set of generated samples and ground truth, motivated by FID~\cite{heusel2017gans}.
To achieve this, we train an MLP-based classifier network with the modified CHI3D dataset to estimate interaction label from given partner and human SMPL-X parameters.
The FHID metric is measured with the MLP feature of the network and this network is also used for top-1 accuracy metrics.

For physical plausibility, we adopt the contact score based on the Chamfer distance and non-collision score.
In the case of contact score, we count body part-wise contact frequencies of interactive humans to calculate potential contact regions corresponding to each interaction label.
Then, the contact score is computed by unidirectional Chamfer distance from interactive human to partner human.
Additionally, the non-collision score, which refers to the ratio of vertices with positive SDF values for the entire vertex of the interactive human is measured following~\cite{Zhang_2020_CVPR, Huang_2023_CVPR}.
%

\begin{table*}[t]
\centering
\resizebox{0.955\linewidth}{!}{%
\begin{tabular}{c|cccc|cccc|cccc} \toprule
\multirow{2}{*}[-2pt]{Interaction} & \multicolumn{4}{c}{IMM} & \multicolumn{4}{c}{SAGA} & \multicolumn{4}{c}{Ours}\\
& FHID & top-1 & contact & non-col & FHID & top-1 & contact & non-col & FHID$\downarrow$ & top-1$\uparrow$ & contact$\downarrow$ & non-col$\uparrow$ \\
\midrule
Push          & 51.00 & 0.00  & 17.52 & 94.08 & \textbf{34.54} & 18.43 & 7.44 & \textbf{95.10} &  56.78 & \textbf{87.52} & \textbf{3.61} & 94.77 \\
Posing        & 174.78 & 11.90 & 14.81 & \textbf{93.54} & \textbf{53.23} & 62.50 & \textbf{4.33} & 92.15 &  71.70 & \textbf{84.15} & 8.58 & 92.72 \\
Grab          & 59.23 & \textbf{92.78} & 13.26 & 96.98 & 46.70 & 84.41 & 5.95 & \textbf{97.00} & \textbf{28.88} & 91.32 & \textbf{4.76} & 96.50 \\
Hug           & 235.84 & 18.72 & 12.45 & \textbf{88.17}  & 57.16 & 93.38 & \textbf{3.16} & 76.21 & \textbf{21.43} & \textbf{99.75} & 5.96 & 78.71 \\
Kick          & 29.89 & 66.67 & 10.01 & \textbf{99.70} & \textbf{42.04} & 30.17 & \textbf{4.62} & 97.41 &  46.55 & \textbf{95.69} & 11.98 & 98.49 \\
Handshake     & 146.90 & 11.90 & 16.14 & 99.10 & 34.39 & 66.12 & \textbf{4.14} & 98.16 & \textbf{12.97} & \textbf{92.12} & 7.34 & \textbf{99.13} \\
Hit           & 43.01 & 61.40 & 10.94 & \textbf{98.52}  & 44.10 & 50.60 & 6.22 & 96.98 & \textbf{20.38} & \textbf{67.47} & \textbf{5.16} & 96.55 \\
Holding Hands & 499.84  & 0.00 & 15.17 & 96.80 & 54.07 & 73.93 & \textbf{4.27} & 96.41 & \textbf{26.20} & \textbf{99.02} & 6.42 & \textbf{97.12} \\
\midrule
All           & 92.25  & 34.49 & 14.24 & 94.41 & 24.26 & 67.47 & \textbf{5.23} & 94.62 & \textbf{6.15} & \textbf{90.30} & 6.07 & \textbf{94.89} \\
\bottomrule
\end{tabular}
}
\vspace{-2mm}
\caption{
    Quantitative comparison on the modified CHI3D dataset.
    \vspace{-4mm}
}
\label{tab.quantitative}
\end{table*}

\begin{table}[!]
\centering
\resizebox{0.9\linewidth}{!}{%
\begin{tabular}{c|cccc} \toprule
    Method & FHID & top-1 & contact & non-col \\
    \midrule
    w/o contact obj. & 6.91 & \textbf{90.70} & 15.11 & \textbf{96.48} \\
    \midrule
      w/ contact obj.       & \textbf{6.15} & 90.30 & \textbf{6.07} & 94.89 \\
    \bottomrule
\end{tabular}
}
\vspace{-2mm}
\caption{
    Ablation study for the interactive optimization.
    \vspace{-4mm}
}
\label{tab.ablation}
\end{table}

\subsection{Comparison Methods} 
Since there are no exact methods that generate an interactive 3D human at the moment of physical contact, we modify and re-implement two related methods for comparisons.
We here briefly discuss how we modify them at a high level. 

\subsubsection{Interaction Mix and Match~\cite{goel2022interaction}.} \
Interaction Mix and Match (IMM) synthesizes interactive human motions, not the contact moment of two humans. 
In addition, IMM generates humans in the form of 3D skeletons, not 3D body meshes like SMPL-X~\cite{pavlakos2019expressive}.
For a fair comparison, we thus pick out 15 joints among the body joints of SMPL-X provided by the CHI3D dataset~\cite{fieraru2020three} for compatibility.
We then train IMM on the modified CHI3D dataset and utilize SMPLify~\cite{bogo2016keep} to render the predicted output into the 3D human mesh.
Lastly, we select frames whose distance between two humans is closer than 2cm as interactive 3D humans from the generated motions.

\subsubsection{SAGA~\cite{wu2022saga}.}
Originally, SAGA is designed to generate human motions conditioned on human-object contacts. To this end, SAGA consists of two-stage: 1) human generation focusing on human-object contacts and 2) motion generation conditioned on the generated human. 
For comparison, we adopt and train the first stage of SAGA for our task on the modified CHI3D dataset.

\subsection{Qualitative Evaluation} 
We conduct qualitative experiments on the modified CI3D dataset to compare our method with SAGA and IMM (see \Fref{fig:qualitative}).
%
SAGA and IMM have difficulties generating physically plausible contact and modeling the dynamic nature of various interactions.
For instance, IMM synthesizes a human that has interpenetration with the given partner without pushing.
Unlike comparison methods, \texttt{\small{ContactGen}} generates physically plausible interactive humans while taking diverse poses corresponding to each type of interaction.
%
%

Furthermore, we additionally compare ours with SAGA in terms of contact prediction (see \Fref{fig:contact_red}).
In the case of SAGA, it just estimates the probability of vertex-wise contact, not the corresponding contact vertex between humans.
Thus, SAGA that minimizing distances between entire feasible regions limits its capacity to capture dynamic interactions.
However, \texttt{\small{ContactGen}} can predict more diverse kinematic relations between interacting humans by estimating part-wise information.
%
\subsection{Quantitative Evaluation} 

We quantitatively evaluate comparison methods and ours on various metrics, such as FHID, top-1 accuracy, contact, and non-collision metrics for both per-category and whole test datasets (see \Tref{tab.quantitative}).
\texttt{\small ContactGen} shows better results than comparison methods in most categories and metrics.
Especially in FHID and top-1 accuracy metrics, our method outperforms comparison methods with large margins.
It means that our method can generate high-quality samples compared to comparison methods.
%

%
%

SAGA, in particular, shows the best contact scores due to its post-optimization step encouraging contact. However, it has limitations in extensively modeling human-to-human interaction, a crucial element for generating natural humans based on the partner. This tendency of SAGA is apparent in both top-1 accuracy and FHID. On the other hand, despite slightly lower contact scores of \texttt{\small ContactGen} compared to SAGA, \texttt{\small ContactGen} outperforms all other scores of SAGA, indicating that \texttt{\small ContactGen} generates more plausible humans maintaining similar contact quality to SAGA.
In the case of IMM, it shows comparable non-collision scores but performs worse in all other metrics. This can be attributed to the fact that IMM does not explicitly model close interaction and contact between humans, allowing for the generation of humans who maintain distance from their partners rather than engaging in close interaction with them.

\subsection{Ablation Study} 
We conduct an ablation study to validate the effectiveness of contact objective in the interactive optimization phase.
As shown in \Tref{tab.ablation}, \texttt{\small ContactGen} with or without contact objective shows comparable performance in general (better for FHID, and comparable to accuracy and non-collision).
However, when we measured the contact metric, \texttt{\small ContactGen} with contact objective outperforms one without contact objective with two times better performance.
%
From this result, we deduce that the contact objective strongly drives the predicted contact regions to be close, resulting in high-quality samples that maintain physical contact.

\section{Conclusion}  \label{sec:conclusion}
In this work, we have proposed \texttt{\small ContactGen}, a novel guided DDM for generating interactive 3D humans in terms of physical contact for a given partner human and desired interaction label.
Especially for dealing with dynamic and non-rigid partners according to desired interaction, we propose the contact prediction module that flexibly estimates potential contact regions. 
This contact prediction module provides valid contact regions to optimize valid interaction regions, which leads the guided DDM to consider physical interaction as guidance.
Thus, \texttt{\small ContactGen} offers a promising solution for generating rich and dynamic human interactions, which involve meaningful and accurate points of contact, enhancing the realism and coherence of the generated human interactions.

\section*{Acknowledgements} 
This work was supported by Institute of Information \& communications Technology Planning \& Evaluation~(IITP) grant funded by the Korea government~(MSIT)
(No. 2022-0-00612, Geometric and Physical Commonsense Reasoning based Behavior Intelligence for Embodied AI and No.2020-0-01336, Artificial Intelligence Graduate School Program~(UNIST)), and the Ministry of Science and ICT and NIPA through the HPC Support Project.

\bibliography{references}

\end{document}